\documentclass{SCIS2022}
\usepackage{xspace}
\newcommand{\ie}{{\emph{i.e.}}\xspace}
\newcommand{\etc}{etc}
\usepackage{graphicx}
\usepackage{amsmath}
\usepackage{}

\usepackage{amssymb}
\usepackage{booktabs}
\usepackage{times}
\usepackage{epsfig}
\usepackage{graphicx}

\usepackage{makecell}
\usepackage{multirow}
\usepackage{diagbox}
\usepackage{makecell}
\newcommand{\pparagraph}[1]{\vspace{1pt} \noindent \textbf{#1} }

\begin{document}
\ArticleType{RESEARCH PAPER}
\Year{2022}
\Month{}
\Vol{}
\No{}
\DOI{}
\ArtNo{}
\ReceiveDate{}
\ReviseDate{}
\AcceptDate{}
\OnlineDate{}

\title{Detecting Building Changes with Off-Nadir Aerial Images}{Title keyword 5 for citation Title for citation Title for citation}

\author[1$\dagger$]{Chao Pang}{}
\author[2$\dagger$]{Jiang Wu}{}
\author[3]{Jian Ding}{}
\author[2]{Can Song}{}
\author[1,3]{Gui-Song Xia}{guisong.xia@whu.edu.cn}

\AuthorMark{Author A}

\AuthorCitation{Author A, Author B, Author C, et al}


\address[1]{School of Computer Science, Wuhan University, Wuhan {\rm 430072}, China}
\address[2]{SenseTime Research, Beijing {\rm 100080}, China}
\address[3]{State Key Lab. of LIESMARS,  Wuhan University, Wuhan {\rm 430079}, China}

\abstract{The tilted viewing nature of the off-nadir aerial images brings severe challenges to the building change detection (BCD) problem: the mismatch of the nearby buildings and the semantic ambiguity of the building facades. To tackle these challenges, we present a multi-task guided change detection network model, named as MTGCD-Net. The proposed model approaches the specific BCD problem by designing three auxiliary tasks, including:
(1) a pixel-wise classification task to predict the roofs and facades of buildings; 
(2) an auxiliary task for learning the roof-to-footprint offsets of each building to account for the misalignment between building roof instances; and 
(3) an auxiliary task for learning the identical roof matching flow between bi-temporal aerial images to tackle the building roof mismatch problem.
These auxiliary tasks provide indispensable and complementary building parsing and matching information.
The predictions of the auxiliary tasks are finally fused to the main building change detection branch with a multi-modal distillation module. To train and test models for the BCD problem with off-nadir aerial images, we create a new benchmark dataset, named BANDON\footnote{We release this BANDON dataset at \url{https://github.com/fitzpchao/BANDON} for reproducible research.}. Extensive experiments demonstrate that our model achieves superior performance over the previous state-of-the-art competitors.}


\maketitle

\section{Introduction}
\label{sec:intro}

Given two registered aerial images captured at different times, building change detection (BCD) aims to detect and localize the image regions
where buildings have been added or torn down between flyovers, and finds applications in many real-world scenarios, such as disaster assessment~\cite{fujita2017damage, gupta2019creating}, urban managements~\cite{dueker1972urban,chen2018improving}, and precise map updating~\cite{awrangjeb2015effective}. 

\begin{figure}[!t]
\vspace{-1mm}
\setlength{\abovecaptionskip}{-0cm}  
\begin{center}
   \includegraphics[width=.92\linewidth]{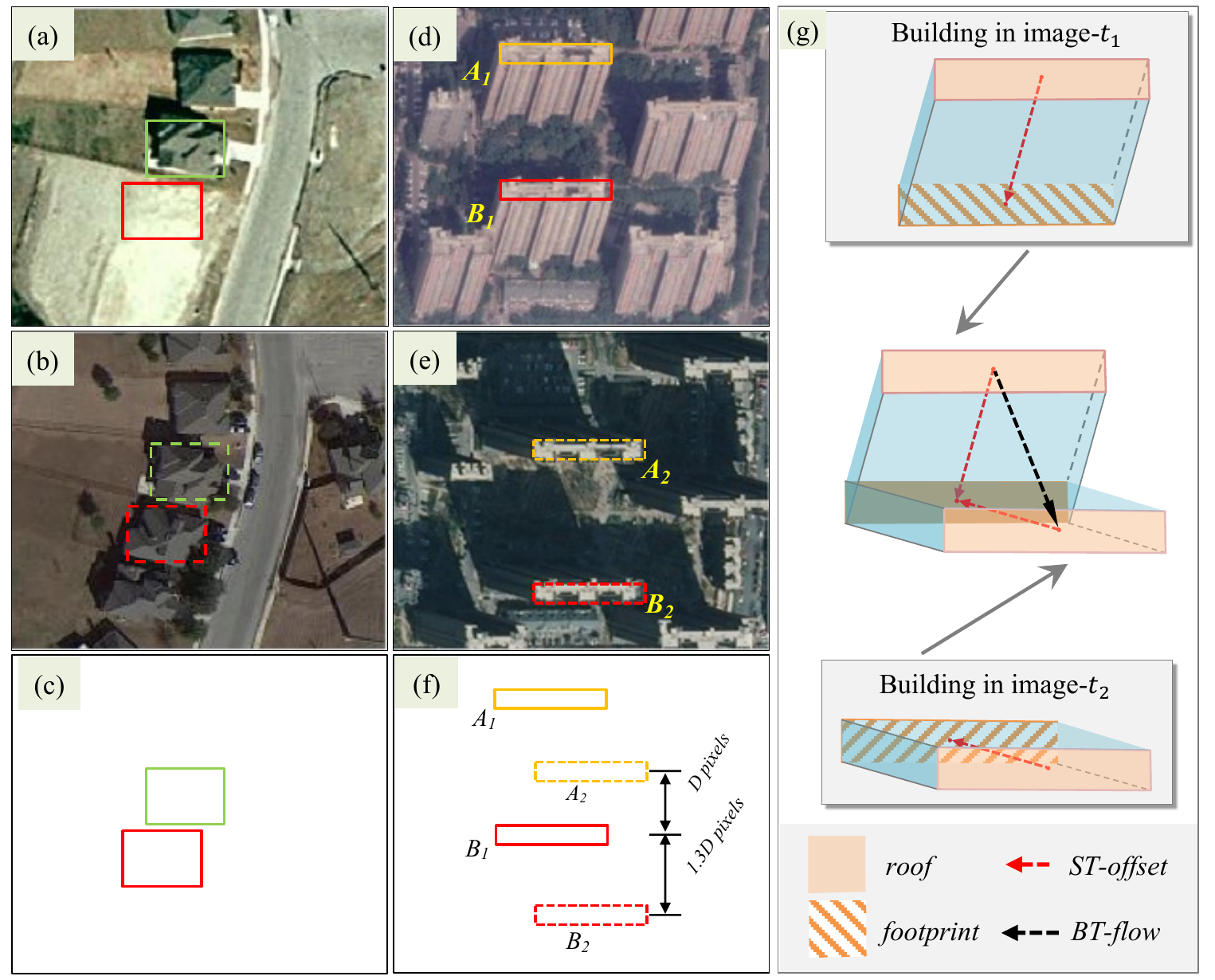}
\end{center}
   \caption{{\bf Building change detection with off-nadir aerial image pairs}. Unlike in near-nadir aerial images (a), the differences between the viewpoints of the two images are non-negligible, and often result in semantic ambiguity to the BCD problem with \textit{off-nadir} images, illustrated by (d)-(e).
   We address these issues by relying on a {\em multi-task guided change detection network}, named MTGCD-Net, which learns the single-temporal building roof-to-footprint offsets (ST-offsets) and the bi-temporal matching flows between identical roofs (BT-flows), shown in (g), to alleviate the semantic ambiguity in the BCD process.
}
\label{fig:1}
\vspace{-2mm}
\end{figure}

The past decade has witnessed promising progress on this change detection task~\cite{nie2021sar} and other visual tasks~\cite{Ding_2019_CVPR, xu2022accurate, ding2012object, kang2017networked, liu2021surface, xu2020special} in remote sensing , by benefiting from the strong representative capability of deep neural networks~\cite{daudt2018fully,chen2021efficient}. However, most of these studies tackle the BCD problem with \textit{near-nadir} images, which are taken nearly from 
a directly top-down view by aerial vehicles or satellites~\cite{chen2020spatial,ji2018fully}. In such a case, the roof and the footprint of buildings in the bi-temporal images are well overlapped, as illustrated by Fig.~\ref{fig:1} (a) and (b). 
Thus, the changed buildings can be easily distinguished by directly comparing the locations of building instances in the image pair. 
Most of the current works~\cite{ji2018fully,sun2020fine} are based on the above observation and with an assumption that the pair of images are pixel-wise aligned.

However, with the popularization of unmanned aerial vehicles (UAVs) and the dramatically increase of tall buildings in cities, it is highly demanded to investigate the BCD problem with \textit{off-nadir} images. As shown in Fig.~\ref{fig:1} (d) and (e), these images are taken at tilted viewing angles.
For tall buildings, the tilted viewing angles lead to significant appearance differences compared with the near-nadir ones: there are non-negligible offsets between the roof and the footprint, and the building facades are also partially visible. 

The non-negligible difference between the viewpoints of bi-temporal images often results in semantic ambiguity to the BCD problem with \textit{off-nadir} images. 
For example, given a certain building instance in one image, one is apt to choose its nearest building with the same roof contours as the counterpart in the other image. However, since the buildings in residential areas usually have similar shape and size, confusions often arise from the spatial misalignment if one only focuses on the building roof. For instance, as shown in Fig.~\ref{fig:1} (d)-(f), $(A_1, \, B_1)$ and $(A_2, \, B_2)$ are the projected contours of the building A and B in the image-$t_1$ and image-$t_2$ respectively. However, since the distance between $(B_1, A_2)$ is smaller than that between $(B_1, B_2)$, $B_1$ is often easily matched to $A_2$ mistakenly and results in detection errors. Another kind of ambiguity comes from the confusion between the building facades and roofs due to the similar shape, texture, and color, which often leads to imprecise change masks. In summary, the appearance differences of tall buildings under different viewing angles introduce semantic ambiguity to the BCD problem with off-nadir images, which makes it significantly more complicated than the near-nadir ones. Thus, it is demanded to design a method specifically for the off-nadir BCD problem.

In this paper, we address the important but under-explored BCD problem with off-nadir images. More precisely, we propose a {\em multi-task guided change detection network}, named MTGCD-Net, by relying on multi-task learning to tackle the off-nadir BCD problem in an end-to-end manner. 

Considering that the single-temporal building roof-to-footprint offsets (ST-offsets)~\cite{li20213d,wang2022pami} and the bi-temporal matching flows between identical roofs (BT-flows), shown in Fig.~\ref{fig:1} (g), can help to alleviate the semantic ambiguity in the BCD process, we approach the problem by designing three auxiliary tasks: (1) \emph{an auxiliary task for learning the pixel-wise classification} to predict building roofs and facades; 
(2) \emph{an auxiliary task for learning the ST-offsets} to relieve the misalignment between building roof instances; and 
(3) \emph{an auxiliary task for learning the BT-flows} to predict the parallax fields of the identical roof between the two images, which can tackle the building roof mismatch problem.
These auxiliary tasks are designed to provide indispensable and complementary information, which are passed to the main BCD branch of the network before the final prediction. Additionally, we also design a multi-task feature guidance module (MTFGM) to enhance the change features for the building change detection task, which extracts and integrates valuable intermediate features from the above auxiliary tasks relying on an attention mechanism. Compared with previous methods, the proposed MTGCD-Net can alleviate the semantic ambiguity in the BCD problem with off-nadir images.


To better train and evaluate our proposed model, we create a new dataset, \ie, \emph{{\bf b}uilding ch{\bf an}ge {\bf d}etection with {\bf o}ff-{\bf n}adir aerial images dataset}, dubbed BANDON, which is composed of off-nadir image pairs of urban and rural areas. Overall, the BANDON dataset contains $2283$ pairs of images, $2283$ change labels, $1891$ BT-flows labels, $1891$ pairs of segmentation labels, and $1891$ pairs of ST-offsets labels ( test sets do not provide auxiliary annotations ), which is superior to many existing building change detection datasets in different aspects, as summarized in Table~\ref{tab:datasets}. So, the BANDON can provide data support for the research of BCD. We evaluate the proposed MTGCD-Net through extensive experiments on the new BANDON benchmark dataset and the SECOND dataset, which demonstrate that our method can largely improve the performance of BCD.

Our main contributions in this paper can be summarized of as follows:
\begin{itemize}
    \item We present a dedicated end-to-end method for addressing the problem of building change detection with {\em off-nadir images}, which explicitly takes into account the semantic ambiguity caused by the difference in viewing angles of tall buildings. 
    \item We design a new model, \ie, MTGCD-Net, relying on a multi-task learning scheme to tackle the challenging off-nadir building change detection task. The carefully designed auxiliary tasks provide indispensable and complementary information, which dramatically helps to improve the main building change detection task.
    \item We create a novel dataset for facilitating the off-nadir building change detection task, \ie, the BANDON. Four types of detailed annotations in the dataset can be used for multi-task learning in aerial images.
\end{itemize}
\section{Related work}

\begin{table}[!t]
\footnotesize
\vspace{-0mm}
\centering
\caption{Comparing different building change detection datasets. N.D. is short for {\em natural disaster}, H.A. is short for {\em human activity}, and $*$ indicates a sequence of $24$ images. }
\vspace{-0mm}
\setlength\tabcolsep{15pt}
\begin{tabular} {ccccccc}
\hline
Dataset & \makecell[c]{$\#$images\\ pairs} & \makecell[c]{image\\size}  & \makecell[c]{change\\type} & \makecell[c]{$\#$change\\ instances} &  \makecell[c]{$\#$change\\ pixels }  \\ \hline 
ABCD\cite{fujita2017damage} &   \makecell[c]{8506\\8444} &  \makecell[c]{160$\times$160\\120$\times$120}  & N.D. & - & - \\
xBD\cite{gupta2019creating} & 22,068 & 1024$\times$1024  & N.D. & - & -  \\
MUDS\cite{van2021multi} & $101^*$ &  1024$\times$1024  & H.A. & - & -  \\ 
SECOND\cite{yang2021asymmetric} & 4662 & 512$\times$512  & H.A. & 59K & 138M   \\
S2Looking~\cite{shen2021s2looking} &5000 & 1024$\times$1024 & H.A. & 66K & 69M \\ 
LEVIR-CD\cite{chen2020spatial} & 637 & 1024$\times$1024  & H.A. & 31K & 30M  \\
WHU-CD\cite{ji2018fully} & 1 & 32,207$\times$15,354  &  H.A. & 2K & 21M  \\ \hline
BANDON (ours) & 2283 & 2048$\times$2048 &  H.A. & 123K & 283M \\
\hline
\end{tabular}

\label{tab:datasets}
\vspace{-0mm}
\end{table}

\subsection{Change Detection Algorithms}
Early works~\cite{dellinger2014change,padron2019kernel} of automatic modeling of building change detection depend largely on hand-crafted features. Recently, a large number of deep models relying on neural networks have been introduced into building change detection ~\cite{bu2020mask,chen2016building}. Among them, fully convolutional siamese structure~\cite{bertinetto2016fully} is the most widely used. FC-Siam-Conc~\cite{daudt2018fully} is a classic change detection model that adopts the siamese fully convolutional neural network structure. It uses two convolutional encoders that share weights to extract two temporal image features. Then the two temporal features are concatenated and sent to a single convolutional decoder to calculate a change mask. 

Subsequent works mainly follow this structure, and advance building change detection in two aspects: 1) designing a more efficient method for feature comparison~\cite{liu2020building,zhang2020deeply,chen2021efficient}, and 2) exploiting joint segmentation tasks~\cite{daudt2019multitask,yang2021asymmetric}. In order to better compare features, the STANet~\cite{chen2020spatial} uses a channel attention module and a spatial attention module to improve the discriminability of changed features. 
Since building change detection is highly related to the semantic segmentation of buildings, some works design networks to jointly learn these two tasks. For example, Ji {\em et al.}~\cite{ji2018fully} exploited bi-temporal building segmentation to directly predict the building change mask. However, this method cannot detect the changes between an old building and a newly reconstructed one, which severely limits the scope of its applications. Liu {\em et al.}~\cite{liu2020building} proposed the DTCDSCN model to learn semantic segmentation and change detection of buildings simultaneously and found that the auxiliary task of building semantic segmentation can improve the performance of the building change detection. However, it only predicted the roofs of buildings in the segmentation. In contrast, we predict both the roofs and facades. Despite the above progress, almost all algorithms are designed and evaluated on building change detection datasets composed of near-nadir images, while most available image pairs are off-nadir. Esfandiari {\em et al.}~\cite{esfandiari2021building} used patch-wise co-registration with the digital surface model for image registration, then used Mask R-CNN~\cite{he2017mask} for building detection, and finally performed a simple comparison to obtain a change mask. This method requires additional data preprocessing, which leads to error accumulation. So, the existing change detection methods are greatly hindered in the application of BCD with off-nadir images, which reflects the necessity of our proposed end-to-end method.

\subsection{Building Change Detection Datasets}

High-quality datasets are the basis for the development and evaluation of building change detection algorithms. Unlike tasks such as semantic segmentation and object detection, datasets related to BCD are relatively rare. In Tab.~\ref{tab:datasets}, we summarize the public BCD datasets. Ji {\em et al.}~\cite{ji2018fully} created the WHU Building Change Detection Dataset (WHU-CD), which only contains a pair of aerial images located in New Zealand. In addition to change detection annotations, it also provides building segmentation annotations, which inspired some further study~\cite{liu2020building} based on multi-tasking learning. Chen {\em et al.}~\cite{chen2020spatial} constructed the LEVIR-CD dataset, which contains 637 image pairs derived from Google Earth. Since the annotations of the above two datasets are well prepared, many existing studies~\cite{zheng2021change,zhang2020deeply} are conducted on the LEVIR-CD and WHU-CD dataset. SECOND ~\cite{yang2021asymmetric} dataset targets to semantic change detection in aerial images, and also contains well-prepared building-related change annotations. MUDS~\cite{van2021multi} is a dataset composed of medium-resolution sequence images and used for building tracking. The ABCD~\cite{fujita2017damage} dataset is composed of images before and after the tsunami and only has image-level annotations. xBD ~\cite{gupta2019creating} is currently the largest public dataset for building change detection, which consists of images of buildings around the world before and after disasters. In addition to change detection and annotation, it also provides natural disaster types and prior image building annotations.
As shown in Tab.~\ref{tab:datasets}, although the scale of xBD and MUDS is large, xBD is mainly dedicated to natural disaster identification and building damage classification, and the low-resolution images of MUDS cannot undertake key tasks such as urban violation detection. Most of the existing datasets are small in scale and are constructed with near-nadir images. Only S2Looking~\cite{shen2021s2looking} dataset contains off-nadir images, and mainly focuses on rural construction. However, urban high-rise buildings are most seriously affected by the imaging angle. So far, there is a lack of a building change detection public dataset with off-nadir city images.

\section{Methodology}

\begin{figure*}[!t]
\vspace{-3mm}
\begin{center}
  \includegraphics[width=\linewidth]{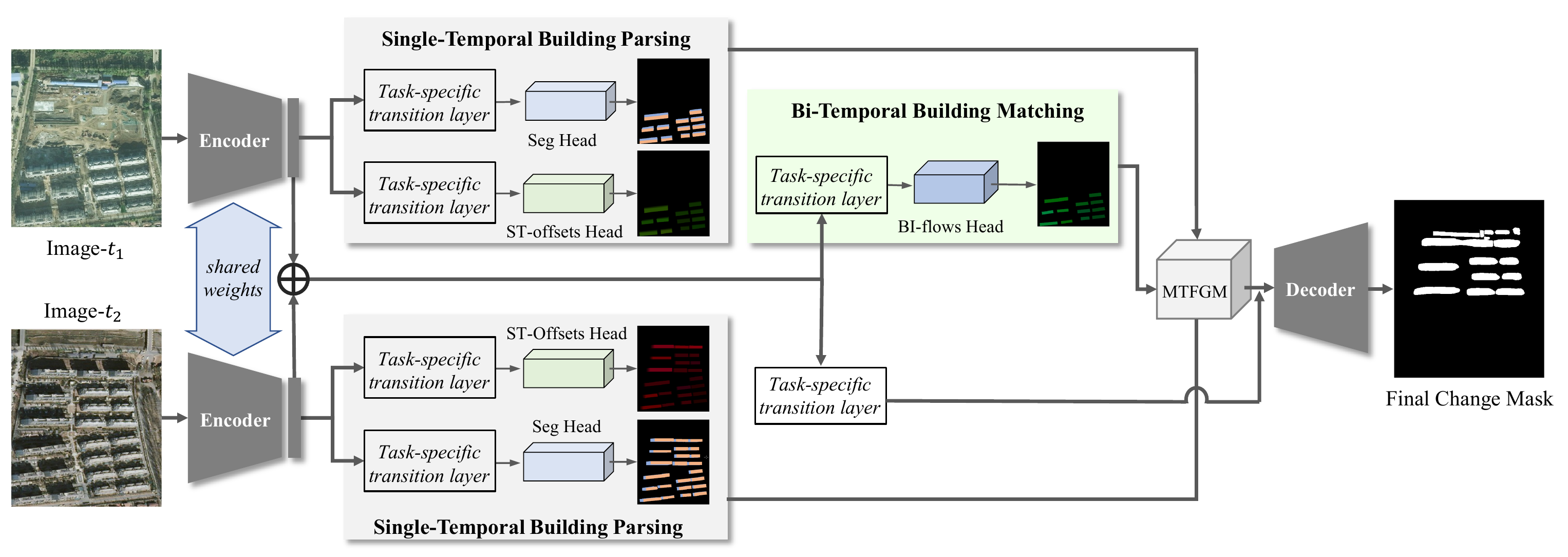}
\end{center}
\vspace{-4mm}
   \caption{Structure of MTGCD-Net, which consists of 3 stages: Firstly, two encoders with the shared weights are used to extract deep features from bitemporal images respectively. Secondly, the deep features of the previous component get multiple intermediate auxiliary task predictions through their respective head. Finally, the intermediate auxiliary tasks assist the prediction of final change detection with a multi-task feature guidance module ( MTFGM ).}
\label{fig:model}
\vspace{-4mm}
\end{figure*}

We first recall the building change detection problem and describe the auxiliary tasks, including the semantic segmentation, the ST-offsets prediction, and the BT-flows prediction. We then present our MTGCD-Net to fuse information from these tasks for building change detection.


\subsection{Change Detection and Auxiliary Tasks}
\label{sub:aux-task}
\pparagraph{Building change detection (BCD).} The BCD takes two bi-temporal images ($I_{t1},I_{t2} \in \mathbb{R}^{ H \times W}$) as inputs, and predicts the area of the changed buildings, which can be represented by a binary mask $Y_{cd} \in \{0, 1\}^{H \times W}$, where $H \times W$ is the size of input images. 

\pparagraph{Auxiliary tasks.} In order to tackle the challenging building change detection on off-nadir image pairs, we define a set of intermediate auxiliary tasks related to BCD. These tasks can be divided into two categories: (1) the \textit{single temporal} building parsing. (2) the \textit{bitemporal} building matching. The \textit{single temporal} building parsing includes the \textit{building semantic segmentation} task and the \textit{ST-offsets prediction} task. The target of the building semantic segmentation task is to predict a segmentation map $Y_{seg} \in \mathbb{R}^{3 \times H \times W}$ with three categories, \ie, background, roofs and facades. The target of the ST-offsets prediction task is to predict a offset $\Vec{V}_{st}=[V_x,V_y]$ for every pixel, where $V_x \in \mathbb{R}^{H \times W}$ and $V_y \in \mathbb{R}^{H \times W}$ are the components of offsets in the horizontal $x$ and vertical $y$ directions. $\Vec{V}_{st}$ is the vector of offset from the roof to footprint of a building in single temporal images (the \textit{red dashed arrow} in Fig.~\ref{fig:1} (g)). The target of bitemporal building matching is to predict $\Vec{V}_{bt}$ for every pixel, which is similar to $\Vec{V}_{st}$ and both are vector, and $\Vec{V}_{bt}$ is the flow vector of identical roof matching with bitemporal images (the \textit{black dashed arrow} in Fig.~\ref{fig:1} (g)). 

\pparagraph{Classification formulation of vector fields prediction.}
The ST-offsets and BT-flows prediction tasks are both \textit{regression} problems, which are difficult to learn during training. 
So our method formulates the two regression as \textit{dense classification} tasks. Based on the distribution statistics of $V_x$ and $V_y$ in these two vector fields, we empirically divide all $V_x$ and $V_y$ into 10 categories.
The ST-offsets and BT-flows finally provide two single temporal offset category map $ Y_{st}^i \in \mathbb{R}^{10 \times H \times W}$ and a bi-temporal matching flows category map $ Y_{bt}^i \in \mathbb{R}^{10 \times H \times W}$, with $i\in \lbrace x, \,y \rbrace$ and $x$ and $y$ being the horizontal and vertical directions respectively.

\subsection{MTGCD-Net}
\label{sub:MTGCD-Net}

Fig.~\ref{fig:model} shows the pipeline of the proposed MTGCD-Net, which consists of 3 stages. Firstly, two encoders with the shared weights are used to extract deep features from bi-temporal images, respectively. Secondly, the deep features are then fed to three parallel heads to obtain the results of the intermediate auxiliary task . Then the features of three tasks  and the original deep features from encoders are combined to obtain the final fine building change detection results with a multi-task feature guidance  module ( MTFGM ). 

\pparagraph{Feature extraction from the bitemporal images.} 
This module takes two bi-temporal images of size $H \times W$ as input. 
Two encoders with shared weights are used to extract the features $ {F}_{ {sh\_ti}} \in \mathbb{R}^{C_{sh} \times \frac{H}{S} \times \frac{W}{S}}$,  $ {i} \in \{0, 1\}, $where $C_{sh}$ is the number of channels, $S$ is the stride of feature maps, and ${i}$ is the time sequence number ( For simplicity, this symbol is generally ignored in subsequent descriptions) . $C_{sh}$ and $S$ are determined by the backbone. 

\pparagraph{Intermediate auxiliary task prediction.} 
Firstly, the features $F_{sh}$ are used to obtain their corresponding single temporal building parsing results, which consists of two parts: the semantic segmentation map $\hat{Y}_{seg} \in \mathbb{R}^{3 \times H \times W} $, and the ST-offsets map $ \hat{Y}_{st} \in \mathbb{R}^{10 \times H \times W}$ . The shared-features are sent to the Task-Specific Transition Layer (TSTL) to get the specific task features $ F_{seg} \in \mathbb{R}^{C_{seg} \times {h} \times{w}}$,  $ {h}=\frac{H}{S}$, $w=\frac{W}{S}$ and $ F_{st} \in \mathbb{R}^{C_{st} \times {h} \times {w}}$ . TSTL is consist of  a convolutional layer, a normalization layer and a nonlinear activation layer connected in series. The number of output feature channels  for each TSTL is set to 512. The specific features are mapped to segmentation masks and offset masks through the corresponding head. 
Secondly, for the bi-temporal building matching task, we first obtain the concatenation of the two features, denoted as bi-temporal features $ F_{cat} \in \mathbb{R}^{C_{cat} \times{h} \times {w}}$, where $ C_{cat}$ is twice as large as $ C_{sh}$ . Then the bi-temporal features are sent to the TSTL and the head to generate a BT-flow feature $F_{bt} \in \mathbb{R}^{C_{cat} \times {h} \times {w}}$, and BI-offsets map $ \hat{Y}_{bt} \in \mathbb{R}^{10 \times H \times W}$, sequentially.  All the above mentioned heads use an FCN-like structure~\cite{long2015fully}.

\begin{figure*}[!t]
\begin{center}
\includegraphics[width=0.8\linewidth]{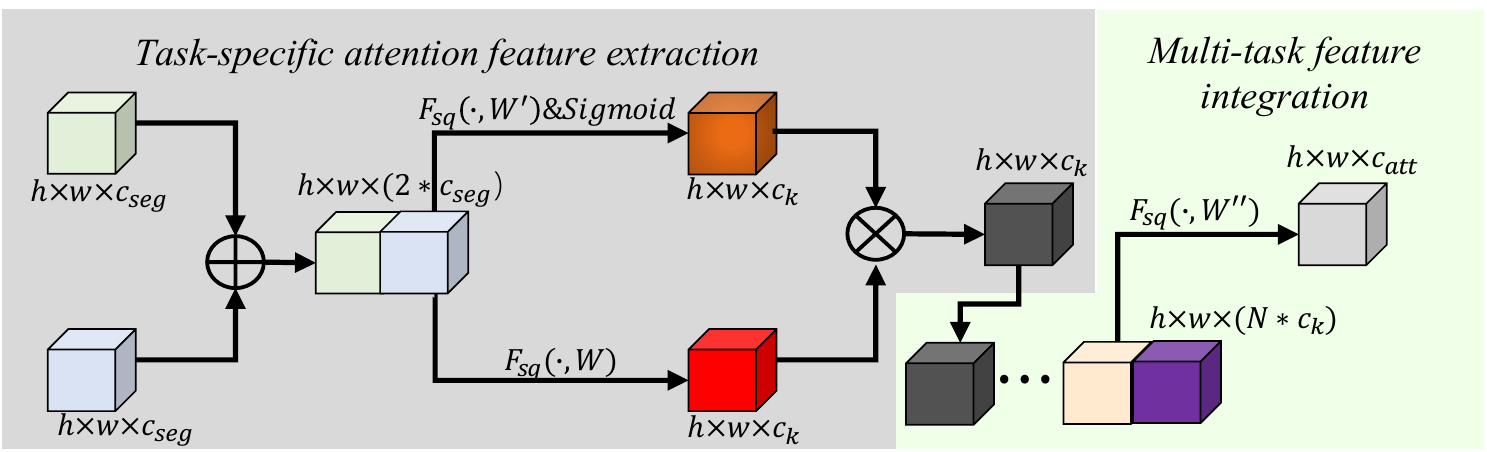}
\end{center}
\vspace{-4mm}
   \caption{Structure of MTFGM, which consists of task-specifc attention feature extraction and multi-task feature integration.}
\label{fig:mtgfm}
\vspace{-1mm}
\end{figure*}

\pparagraph{ Multi-task feature guidance module.}  To obtain valuable information from  auxiliary task features, we propose a MTFGM, which consists of two parts: task-specifc attention feature extraction and multi-task feature integration.  The details of the  are shown in Fig.~\ref{fig:mtgfm}. MTFGM  tasks all feature , \ie, $ F_{seg}$, $ {F}_{st}$ and $F_{bt}$,  from auxiliary tasks as inputs.  With the built-in attetion layer, this module will obtains  features that can guide the learning of the change detection task. \\
In the task-specifc attention feature extraction part, attention feature extraction is performed on the three types of input respectively. For simplicity,   we only present an illustration of one type of input in the task-specifc attention feature extraction part of Fig.~\ref{fig:mtgfm}.  We take the segmentation feature input branch as an example to introduce this part of the network structure. We first obtain the concatenation of the two segmentation features, denoted as bi-temporal segmentation features $ {F}_{seg}^{bi} \in \mathbb{R}^{2*C_{seg} \times {h} \times {w}}$.  Then $ {F}_{seg}^{bi}$ is passed to two different squeeze layers $\mathcal{F}_{sq} $ to get attetion weight feature  $ {F}_{seg}^{wgt} \in  {[0,1]}^{C_{k} \times {h} \times {w}}$  and reduced feature  $ {F}_{seg}^{'} \in \mathbb{R}^{C_{k} \times {h} \times {w}}$ , respectively. Finally, multiply $ {F}_{seg}^{wgt} $ and $ {F}_{seg}^{'}$ to get the final segmentation task attention feature  ${F}_{seg}^{att} \in \mathbb{R}^{C_{k} \times {h} \times {w}}$. This calculation process can be denoted as

\begin{equation}
    {F}_{seg}^{'} =  \mathcal{F}_{eq} ({F}_{seg}^{bi}, W)), \\
    {F}_{seg}^{wgt} =  Sigmoid(\mathcal{F}_{eq} ({F}_{seg}^{bt}, W^{'})) , \\
	{F}_{seg}^{att} =  Multiply({F}_{seg}^{'}, {F}_{seg}^{wgt}).
\label{equ:1}
\end{equation}

For features derived from other auxiliary tasks, attention features ${F}_{st}^{att} \in \mathbb{R}^{C_{k} \times {h} \times {w}}$ and ${F}_{bt}^{att} \in \mathbb{R}^{C_{k} \times {h} \times {w}}$ are obtained in the same way.  In the multi-task feature integration part,  all features are fed into a squeeze layer to get the final multi-task guidance feature ${F}_{att}\in \mathbb{R}^{C_{att} \times {h} \times {w}}$ , which can denoted as
\begin{equation}
    {F}_{att} =  \mathcal{F}_{eq} (Concat([F_{seg}^{att}, F_{st}^{att},F_{bt}^{att}]), W^{''})) .
\label{equ:1}
\end{equation}

\pparagraph{ Change detection prediction } The final change detection task also takes the original bi-temporal feature $F_{bi}$  and the multi-task guidance feature ${F}_{att}$ as inputs.  The $F_{bi}$ is also fed into a TSTL to get the original change feature $F_{cd}$. The  $F_{cd}$ canbe used directly to predict the binary change mask, but the result is \textit{coarse} because $F_{cd}$ not explicitly incorporate the effective information from the other tasks. So, the final decoder takes the refined features $F_{fcd}$, which is the concatenation of   ${F}_{att}$ and  $F_{cd}$ to predict the \textit{fine} change mask $\hat{Y}_{cd} \in \{0, 1\} ^ { H \times W} $, and is composed of a TSTL, an up-sampling layer and an FCN-Head. 

 \subsection{The Objective Function}
 \label{sub:loss}
The overall outputs $\Hat{Y}$ of our MTGCD-Net consists of a semantic segmentation map $\hat{Y}_{seg}$, an ST-offsets map $\hat{Y}_{st}$, a BI-flows map $\hat{Y}_{bt}$,  and a  change map $\hat{Y}_{cd}$.
Denoting the loss function of the above tasks by $\mathcal{L}_{seg}$, $\mathcal{L}_{st}$, $\mathcal{L}_{bt}$,  $\mathcal{L}_{cd}$,
the MTGCD-Net is trained by optimizing the overall loss $\mathcal{L}_{tot}(\hat{Y},{Y})$, as 
 \begin{equation}
    \mathcal{L}_{tot} = \mathcal{L}_{cd} + \lambda_{1}\mathcal{L}_{seg} + \lambda_{2} \mathcal{L}_{st} + \lambda_{3} \mathcal{L}_{bt} ,
\label{equ:loss}
\end{equation}
which is a linearly combined optimization objective, with task weightings $\lambda_{i}, i= \{ 1,2,3\}$, and $Y$ is the label of all output predictions $\hat{Y}$. $\mathcal{L}_{cd}$  adopt a combination of Cross Entropy Loss and Dice Loss~\cite{sudre2017generalised} with the same weight, and other loss functions only involve Cross Entropy Loss.

\section{BANDON Dataset}

\begin{figure}[!t]
\vspace{-2mm}
   \centering
   \includegraphics[width=.97\linewidth]{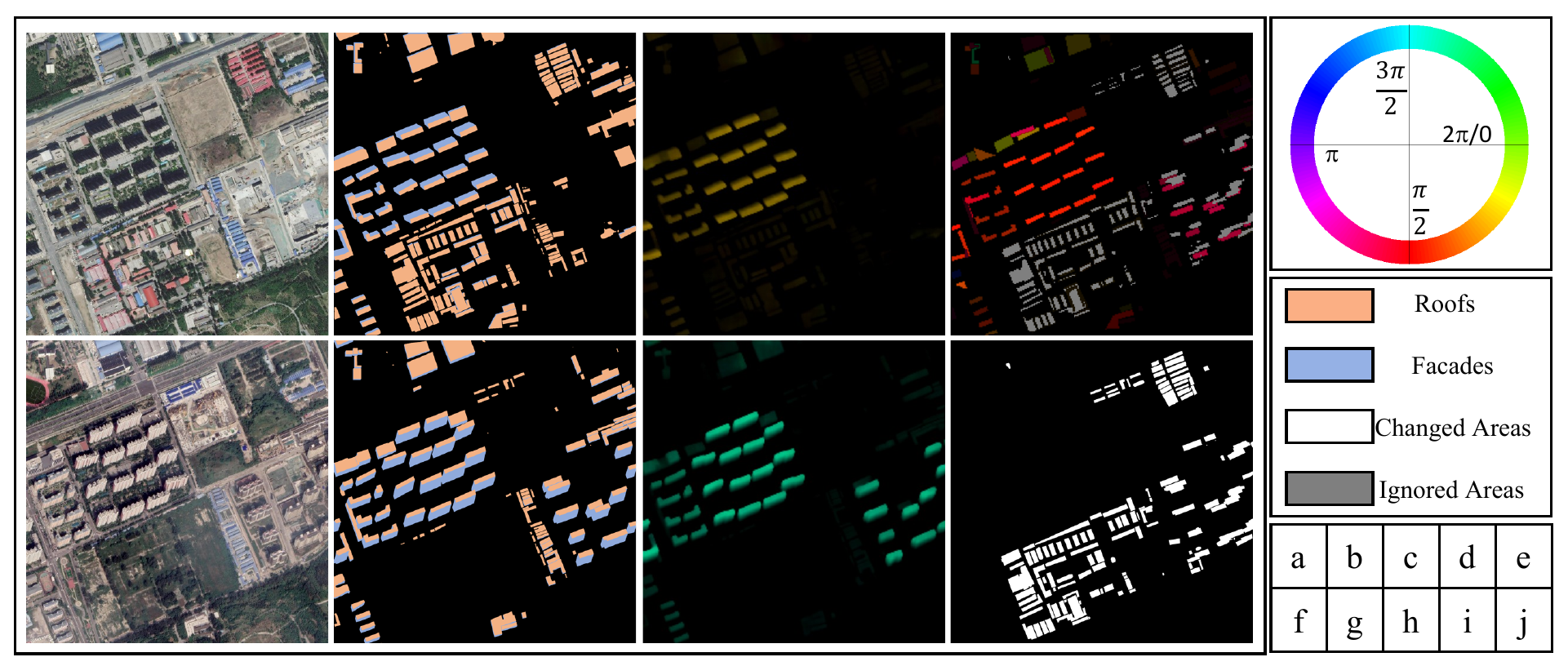}
   \vspace{-2mm}
   \caption{ A sample (a pair of images and their corresponding annotations) from our BANDON dataset. (a) and (f) are the images at $t_1$ and $t_2$, respectively. (b) and (g) are the segmentation labels of the two images. (c) and (h) are the corresponding ST-offsets labels of the images at $t_1$ and $t_2$. (d) is the BT-flows labels between the two images. (j) is the changed mask. (e) is the legend of direction used in (c), (h), and (d), which indicates directions by different colors. (j) is the legend used in (b), (g) and (i). }
\label{fig:dataset}
\vspace{-1mm}
\end{figure}
\subsection{Dataset Overview}

We build the BANDON dataset with a large quantity of off-nadir aerial images with the spatial resolutions of 0.6m, mainly collected from Google Earth images\footnote{\url{https://earth.google.com/}}, Microsoft Virtual Earth images\footnote{\url{http://www.microsoft.com/maps/}} and ArcGIS images\footnote{\url{https://www.arcgis.com/apps/mapviewer/}}. The images are geo-located in  six representative cities of China, \textit{i.e.}, Beijing, Shanghai, Wuhan, Shenzhen, HongKong and Jinan. In order to achieve high generalization performance of the dataset and to be closer to real application scenarios, we select three images of different times and sources for a location to form three pairs of image samples.

\begin{figure}[!t]
\vspace{-2mm}
   \centering
   \includegraphics[width=1.0\linewidth]{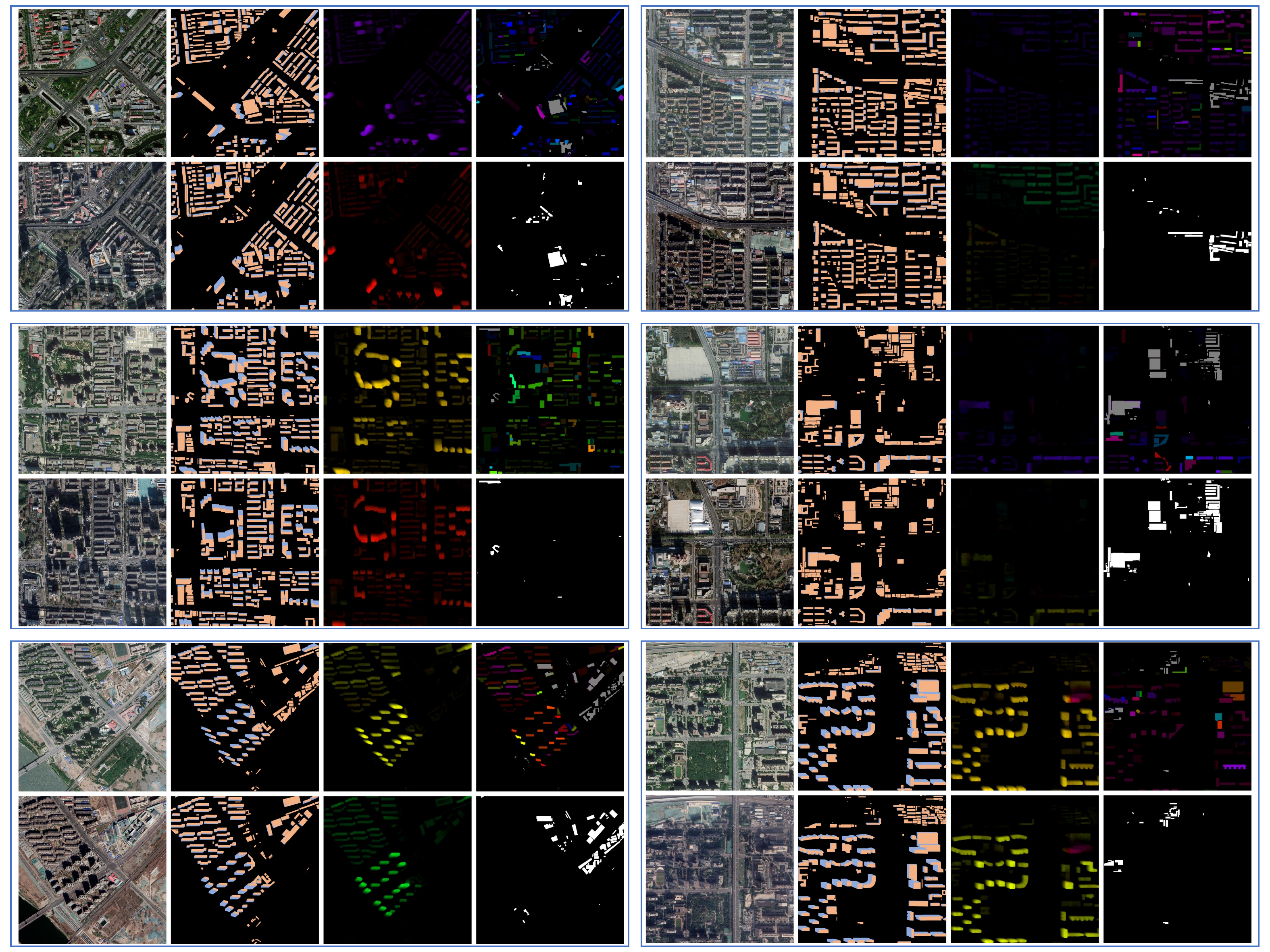}
   \vspace{-2mm}
   \caption{More samples from BANDON. }
\label{fig:more_samples}
\vspace{-3mm}
\end{figure}

We provide holistic annotations to track the BCD problem with off-nadir images. There are four types of annotations provided by BANDON:1) semantic segmentation labels with three categories, (\textit{i.e.}, background, roofs and facades); 2) the labels of single-temporal building roof-to-footprint offsets; 3)  the labels of bi-temporal matching flows between identical roofs 4) building change labels. Fig.~\ref{fig:1}(g) illustrates an example of annotation types containing 1)-3) with an identical building in BANDON dataset. Fig.~\ref{fig:dataset} shows a complete samples of BANDON dataset. To better understand the details of the BANDON dataset, we provide more visualization samples  in Fig.~\ref{fig:more_samples}.

BANDON contains 1689 training image pairs, 202 validation image pairs and 392 test image pairs, and each image is cropped into $ 2048 \times 2048 $ pixels. To better evaluate the generalization capacity of the proposed method, we divide our test set into an in-domain test set containing 207 image pairs located in the same two cities(Shanghai and Beijing) but different regions with the training set, and an out-domain test set containing 185 image pairs located in  four cities (Jinan,Wuhan, Shenzhen and HongKong) that is not included in the training dataset. We will release this BANDON dataset at {\color{red}\emph{https://github.com/fitzpchao/BANDON}} for reproducible research.

\subsection{Annotation Details}

\paragraph{Creating Building Change Labels.}
We adopted a semi-automatic labeling strategy for creating building change labels of training set, validation set and in-domain test set, \ie, an automatic candidate generation step followed by manual annotations. For out-domain test set, we all use manual annotation to provide building change detection labels. \\

\emph{Automatic Candidate Generation.} First, we use the change detection API\footnote{https://rs.sensetime.com/se/index.html\#/} provided by SenseTime to produce general change candidate regions with high recall rate ($>97\%$) for image pairs in BANDON. However, these generated general candidate regions include changes not only buildings but also other objects, such as changes from fields to bare soil, and roads to bare soil,~\etc. Second, for each image in BANDON, we also generate the roof masks of buildings by the same API, with high precision. Subsequently, to get clean building change candidate areas, we use the roof masks to filter out non building changes from the general candidate areas.\\

 \emph{Manual Annotation.} The automatically produced building change candidate regions have not only negligible missed detection, but also a large number of false detection, especially in off-nadir building areas. Thus, we ask remote sensing experts to manually correct these candidate regions, and the automatic generation of candidate regions can greatly speed up the annotation process. After the above operations, the final binary building change detection labels $Y_{cd}$ get prepared.
 
\paragraph{Creating Auxiliary Labels.}
The labels involved in the three types of auxiliary tasks are annotated by three steps. First, for each location, we ask remote-sensing experts to manually annotate an instance-wise building roof map $Y_{roof}^{t_1}$ of the images at $t_1$ with unique ID. 
Second, the annotators are asked to mark the offset vector~\cite{li20213d} from the roof to the footprint for each building on the $Y_{roof}^{t_1}$, and we can subsequently generate the single-temporal building roof-to-footprint offsets (ST-offsets) $Y_{st}^{t_{1}}$ and the semantic segmentation labels $Y_{seg}^{t_{1}}$ with three categories, (\textit{i.e.}, background, roofs and facades) by using the building roof $Y_{roof}^{t_1}$ and the offset vector annotations. For the annotations of the images at $t_2$ date, we create the building footprint map $Y_{foot}^{t_{1}}$ using the shape of roofs in $Y_{roof}^{t_1}$. We use $Y_{foot}^{t_{1}}$ and $Y_{cd}$ to generate the building footprint map $Y_{foot}^{t_{2}}$. Buildings that have changed in $Y_{foot}^{t_{2}}$ are ignored to reduce the labeling cost, as show in gray in Fig.~\ref{fig:1}. In the experiments, the ignored areas do not participate in the loss calculation, and the same ignored areas are also located on the single-temporal building roof-to-footprint offsets (ST-offsets) and the bi-temporal matching flows between identical roofs (BT-flows) label. Then, we can get the semantic segmentation labels $Y_{seg}^{t_{2}}$ and the ST-offsets $Y_{st}^{t_{2}}$ by manually labeling the offset vector from roof to footprint based on $M_{foot}^{t_{2}}$. Final, we use the unique ID of buildings in $Y_{st}^{t_1}$ and $Y_{st}^{t_2}$ to match the identical unchanged building, and based on the same location of the identical building footprint on $Y_{st}^{t_1}$ and $Y_{st}^{t_2}$ , the BT-flows label $Y_{bi}$ can be generated by using $Y_{st}^{t_1}$ and $Y_{st}^{t_2}$.

\section{Experiments}

\subsection{Implementation Details}
\label{sec:exp_details}
We train each model for comparison on our BANDON datasets. All these models are initialized by the weights pretrained on ImageNet. We train these models on 8 GPUs using SGD with the poly decayed learning rate schedule \cite{zhao2017pyramid} starting from 0.01 for 40,000 iterations. The batch size is 16. For the data augmentation, horizontal and vertical flip, rotation (with the angle in $[90, 180, 270, 360]$), and color jittering are used. The images are then randomly cropped into 512$\times$512 patches for training . Since each temporal image has the predictions of the semantic segmentation map and the ST-offset map, we empirically set $\lambda_{1},\lambda_{2}$ , $\lambda_{3}=1$. 

We adopt the PSPNet~\cite{zhao2017pyramid} with the ResNet-50~\cite{he2016deep} backbone as the basic architecture of the MTGCD-Net model, unless specified otherwise. In the experiment, the structure of the MTGCD-Net model without the auxiliary task learning part is used as the baseline model. Considering the diversity of image styles in the BANDON dataset, we replace the model's first batch normalization layer with an instance normalization layer. 
We use foreground Intersection-over-Union (IoU) as our major metric. Also, the typical F1-score ($F_1$), Recall and Precision  measurement on foreground pixels are provided. The foreground here indicates the changed region of buildings in both prediction and ground truth.

\begin{table}[!t]
\vspace{-0mm}
\footnotesize

\begin{center}
\caption{Quantitative comparison with state-of-the-art  methods on the BANDON test dataset. }
\vspace{-0mm}
\setlength\tabcolsep{9pt}
\begin{tabular} {c|c|cccc|cccc}
\hline
\multirow{2}{*}{Method} & \multirow{2}{*}{Backbone} & \multicolumn{4}{c|}{In-domain Test}  & \multicolumn{4}{c}{Out-domain Test} \\ \cline{3-10}
 & & IoU & $F_1$ & Recall & Precision  & IoU & $F_1$ & Recall & Precision\\ \hline
 FC-EF\cite{daudt2018fully} & ResNet-50 &26.30 &41.65& 34.83 & 51.80& 21.24 &35.04 &30.73& 40.76	\\
 FC-siam-Diff\cite{daudt2018fully} & ResNet-50 & 49.51&	66.23&	67.39&	65.11 & 43.06&	60.20&	60.06&	60.33\\
 FC-Siam-Conc\cite{daudt2018fully} & ResNet-50 & 49.06 &	65.83	&66.17 &	65.49 &42.56	&59.71	&58.65	&60.81 \\
 DTCDSCN\cite{liu2020building} & SE-ResNet-34  & 46.05 &63.06 &57.90& 69.23	 &38..43& 55.53& 56.89 &54.22	 \\
 STANet-BAM\cite{chen2020spatial} & ResNet-50  & 51.46	&67.95&	69.44&	6653&44.41&	61.50&	62.07	&60.94 \\
 STANet-PAM\cite{chen2020spatial} & ResNet-50  & 51.28	&67.79	&69.42&	6624&44.01&	61.12&	61.92	&60.35 \\
 Change-Mix\cite{zheng2021change} & ResNet-50 & 53.89     &70.04	&70.80	&69.29  & 47.50 &	 64.41&	63.26 &	65.60 \\ 
ChangeFormer~\cite{bandara2022transformer} & SegFormer-b2 & 52.76 &	69.07	& 72.26	& 66.15 & 45.56 &	62.60 &	64.61	& 60.72 \\
\hline
Baseline & ResNet-50 &  54.14&	70.25&	71.26&	69.26 & 46.91&	63.86&	61.88&	65.98   \\
 Ours & ResNet-50 & \textbf{60.27} & \textbf{75.21} & \textbf{76.55} & \textbf{73.92} & \textbf{53.08} & \textbf{69.35} & \textbf{67.30} & \textbf{71.52} \\ \hline
\end{tabular}
\label{tab:sota}
\end{center}
\vspace{-4mm}
\end{table}

\begin{figure*}[!t]
\vspace{-3.3mm}
\setlength{\abovecaptionskip}{-0.1cm}  
\centering
   \includegraphics[width=.99\linewidth]{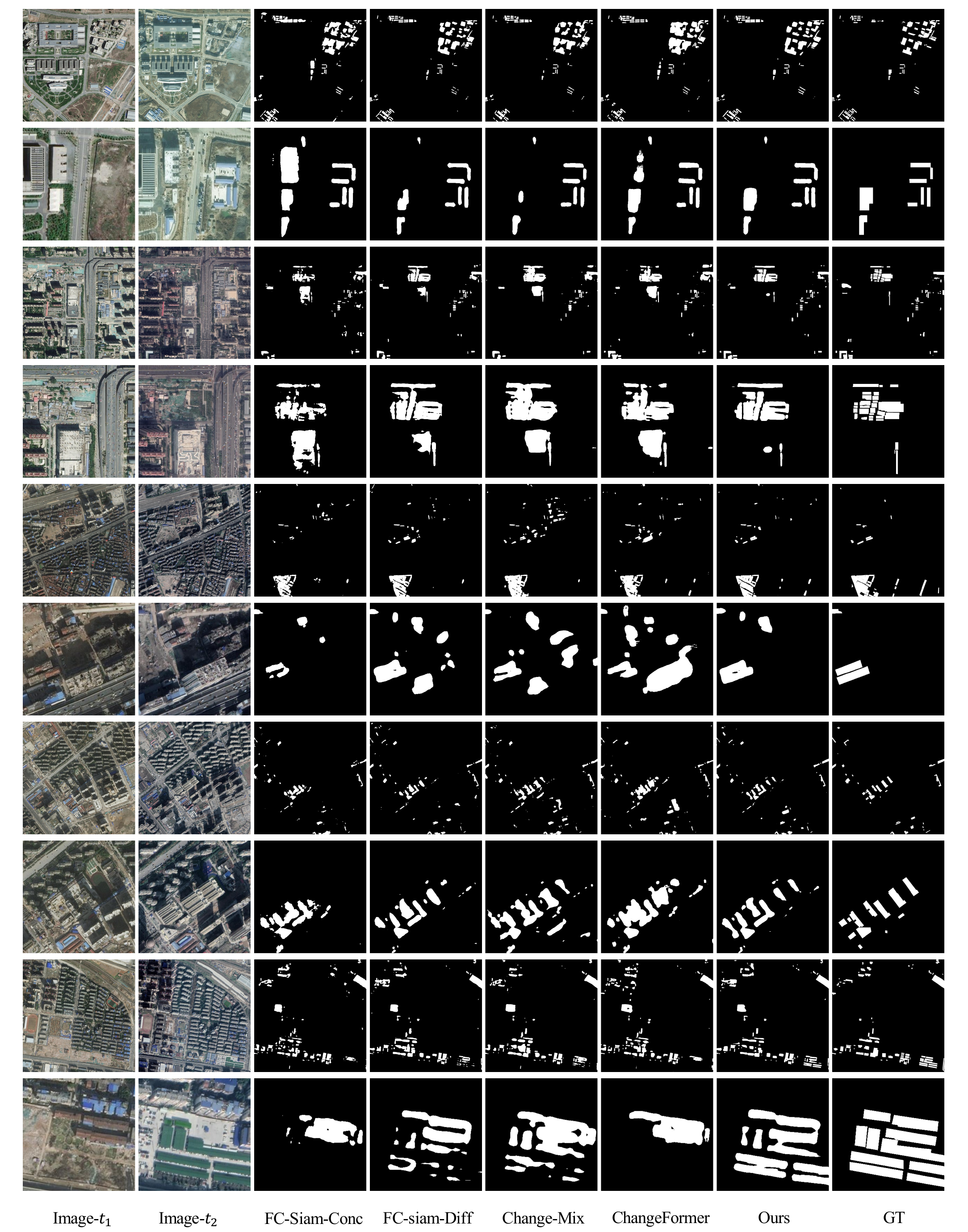}
\vspace{1mm}
\caption{Comparison with the state-of-the-art works.The white represents the changed area. The odd rows show the results for test images with broader coverage. The even rows show the close-up of detection results by different methods, which highlights the better performance of our proposed method.}
\label{fig:sota}
\vspace{-4mm}
\end{figure*}

\subsection{Comparison with the State-of-the-art}
Tab.~\ref{tab:sota} shows that the performance of the various building change detection algorithms on BANDON dataset.  We find that all algorithms achieve poor performance, which proves that the non-orthophoto scenes in the BANDON dataset are challenging for current algorithms. Most algorithms perform as expected, but FC-EF performs extremely poorly, which should be attributed to the large spatial misalignment of the same building in the bitemporal off-nadir image pair and the absence of a pretrained model with 6-channel input image. \\
 The core of our method is to provide more comprehensive information for change detection by introducing auxiliary tasks to obtain better change masks. So, we also compare the performance of our methods with the state-of-the-art methods on BANDON dataset in Tab.~\ref{tab:sota}, without considering whether the method introduced auxiliary annotations during training.  With the learning of auxiliary tasks, MTGCD-Net achieves a better performance than all the other methods on both the in-domain and the out-domain test set in all metrics. By observing Tab.~\ref{tab:sota}, we found that our baseline model has achieved competitive results, surpassed only by the change-mix method on the out-domain test set. This is due to the model architecture and the introduction of instance normalization layer. Our model explicitly extracts valuable information from the intermediate features of several auxiliary tasks and performs feature integration for the change detection task, which is based on the attention mechanism. So, our method improves the performance on two test sets by 6.13$ \% $, 6.17$ \% $ in IoU and 4.96$ \% $, 5.49$ \% $ in ${F}_{1}$. We visualize the building change detection results of our model and some representative methods in Fig.~\ref{fig:sota}. 
  We provide the visualization results at the size of 2048 $\times $ 2048 pixels and their close-ups, corresponding to the odd and even rows in Fig.~\ref{fig:sota}. The close-up can highlight the strengths of our method. As shown in the first four even rows, our model can significantly alleviate the false detection of the off-nadir buildings in various scenarios. As shown in the last even rows, our MTGCD-Net can also improves the ability to detect the building changes in the near-nadir images.
 As in the odd rows, the results of images with a large coverage area can better reflect the strengths and weaknesses of methods. Although the performance of our method has improved compared to others, the results are still very unsatisfactory compared with ground truth, which reflects that building change detection with off-nadir images is a very challenging task. 


\subsection{Ablation Study}
We conduct a series of experiment to investigate the function of each component in the proposed method. The detailed comparisons are given in the following.

\begin{table}[!t]
\footnotesize

\centering
\caption{Ablation studies on the test set of BANDON. Three auxiliary tasks are considered in this ablation study, as {\bf (a)} the semantic segmentation task, {\bf (b)} the ST-offsets prediction task, and {\bf (c)} the BT-flows prediction task. }

\setlength\tabcolsep{17pt}
\begin{tabular} {c|c c c |c c | c c}
\hline
\multirow{2}{*}{Structure} & \multicolumn{3}{c|}{Auxiliary Task} & \multicolumn{2}{c|}{In-domain Test}  &  \multicolumn{2}{c}{Out-domain Test} \\ \cline{2-8}

 &{\bf (a)} &{\bf (b)}  &{\bf (c)}  &IoU & $F_1$  & IoU & $F_1$ \\ \hline

Baseline(BN) & & & & 53.10 &	69.36 & 46.23	 &63.23  \\
Baseline & & & & 54.14	 & 70.25	 & 46.91 &	63.86	 \\
 Baseline & $\checkmark$ & & & 59.15	 &74.33	 & 51.90	 &68.33	 \\
 Baseline   & & $\checkmark$ & & 58.90  &74.14 	& 52.14  &68.54 	 \\ 
 Baseline  & & &$\checkmark$  &57.55 & 73.06 	 & 50.39 & 67.01 	 \\
 \hline
 Baseline  & $\checkmark$ & $\checkmark$ & $\checkmark$ & 58.66&73.95 &51.56&	68.04 \\
 MTGCD-Net  & $\checkmark$ & $\checkmark$ & $\checkmark$ & \textbf{60.27} & \textbf{75.21} & \textbf{53.08} & \textbf{69.35} \\

\hline
\end{tabular}

\label{tab:abl_1}
\vspace{-3mm}
\end{table}
\pparagraph{Influence of the instance normalization layer.} By replacing the batch normalization layer with instance normalization layer, the performance of the model on the two test sets improved by 1.04$\%$, and 0.63$\%$, respectively. The comparison of other metrics is shown in Tab.~\ref{tab:abl_1}. We can note that instance normalization performs style normalization by normalizing feature statistics, which have been found to carry the style information of an image, as in \cite{huang2017arbitrary}. Furthermore, due to the difference in the acquisition time and sensor platforms, there is a non-negligible radiation difference between the two images to be detected. So instance normalization layer can process the styles of the two images to be consistent, and it is more effective to find changes.

\begin{figure*}[!t]
\vspace{-1mm}
\centering
   \includegraphics[width=1.0\linewidth]{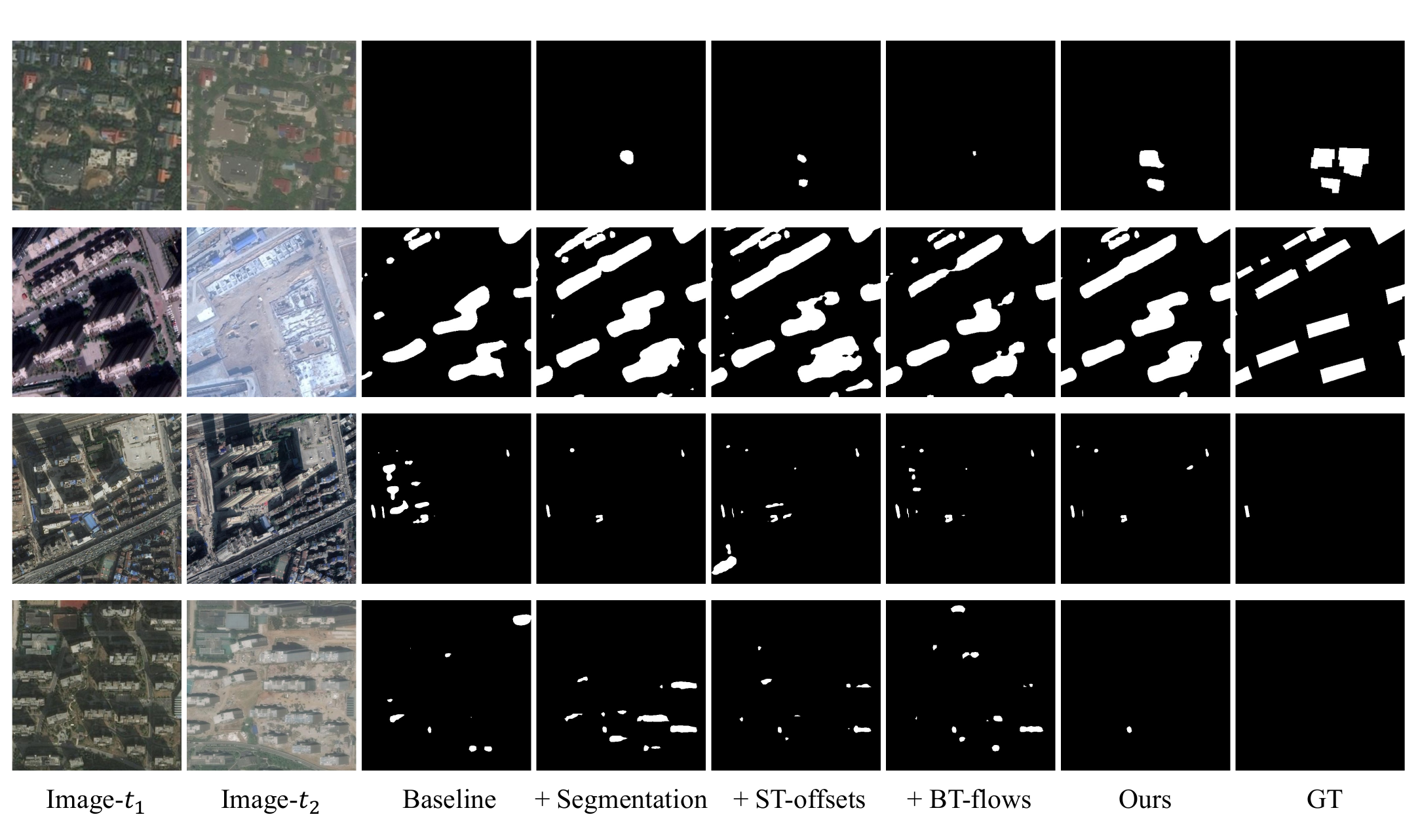}
   \caption{Qualitative results of auxiliary tasks and our MTGCD-Net. The white represents the change area. }
\label{fig:abl}
\vspace{-1mm}
\end{figure*}

\pparagraph{Influence of the auxiliary tasks.} We compared the performance promotion from each of the intermediate auxiliary tasks to the final predictions. We directly add the FCN-Head to the encoder as the decoder to verify the effectiveness of the auxiliary tasks. Tab.\ref{tab:abl_1} shows the comparison results. We can observe that the semantic segmentation task of 3 categories , the ST-offsets prediction task, and the BI-flows prediction task can improve the performances of the model on out-domain test set by 4.99$\%$, 5.23$\%$, 3.48$\%$ in IoU, respectively. Essentially, the auxiliary task of building segmentation provides delicate contour information for the main task of change detection, so the predicted contour is complete. The ST-offsets prediction task provides roof-to-footprint offset information of the building to relieve the misalignment between building roof instances. The BI-flows prediction task provides unchanged building matching information to reduce false detection caused by pixel misalignment of the identical building between the two images. Particularly, the ST-offsets task brings the largest improvement to the model on  out-domain test set ( from 46.91$\%$ to 52.14$\%$ in IoU), which demonstrates the effectiveness of resolving the probable mismatch of nearby buildings due to the spatial misalignment in off-nadir images, as shown in Fig.~\ref{fig:1}. Qualitative examples are shown in Fig.~\ref{fig:abl}.  It can be observed that the additional semantic segmentation task can improve the quality of the change mask contours. Each task can improve the false detection problem of the tall buildings in off-nadir images to a certain extent. \\ 
\pparagraph{Influence of the multi-task feature guidance module.} As shown in Tab.~\ref{tab:abl_1}, when three auxiliary task prediction heads are added at the same time, the performance is not as good as the results of model with a single auxiliary task. Although there are correlations between different tasks, there still are differences. For four different tasks, it is normal that the differences will lead to a decline in the feature extraction ability of the model. So, we propose MTGFM excavate the correlation between auxiliary tasks and change detection tasks, which uses the task-specific feature of each task to explicitly guide the learning of the final change detection task. In detail, the MTGFM explicitly utilizes the intermediate features of auxiliary tasks to mine more valuable information for the change areas prediction. So, the addition of the MTGFM results in about 2.52$\%$ IoU improvement based on the implicit guidance on the out-domain test set. We also visualize part of the experimental results in Fig.~\ref{fig:abl}



\begin{table}[!t]
\footnotesize
\centering
\caption{Comparison of different building segmentation prediction strategies. seg indicates the Building Semantic Segmentation Auxiliary Task.}
\setlength\tabcolsep{15pt}
\begin{tabular} {c|c cc | c c| cc}
\hline
\multirow{2}{*}{ Structure} & \multicolumn{3}{c|}{Semantic categories} & \multicolumn{2}{c|}{In-domain Test}  &  \multicolumn{2}{c}{Out-domain Test} \\ \cline{2-8}
  &{\bf background}  &{\bf roofs} & {\bf facades}  &IoU & $F_1$  & IoU & $F_1$ \\ \hline
Baseline & &  & & 54.14	 & 70.25	 & 46.91 &	63.86	 \\ \hline
Baseline + Seg & 0&1 &0 &57.79	& 73.25	& 50.84	&67.41 \\
Baseline + Seg  & 0&1&1 & 57.92 &	73.35	 &50.39 &	67.01	 \\ 
 Baseline + Seg  &0&1& 2& 59.15	 &74.33	 & 51.90	 &68.33 \\
\hline
\end{tabular}
\label{tab:abl_seg}
\vspace{-3mm}
\end{table}

\pparagraph{Influence of building semantic categories.}
 Moreover, since the previous methods only used the background and the building roof as the two-category semantic segmentation task, so we compared the segmentation tasks with and without the building facades category. As shown as in Tab.~\ref{tab:abl_seg}, the additional building facades category brings the improvement of 1.36$\%$ in IoU on in-domain test set  and 1.06 in $\%$ IoU on out-domain test set, demonstrating the effectiveness of alleviating the semantic ambiguity. In addition, we also find that the two semantic strategies have similar effects on the model when the building facades belong to the background and foreground categories, respectively.

 \begin{table}[!t]
\footnotesize

\centering
\caption{Comparison of different offset/flow field prediction strategies.}

\setlength\tabcolsep{17pt}
\begin{tabular} {c|c c |c c | c c}
\hline
\multirow{2}{*}{Structure} & \multicolumn{2}{c|}{Learning strategies} & \multicolumn{2}{c|}{In-domain Test}  &  \multicolumn{2}{c}{Out-domain Test} \\ \cline{2-7}

  &{\bf regression}  &{\bf classfication}  &IoU & $F_1$  & IoU & $F_1$ \\ \hline
Baseline & &  & 54.14	 & 70.25	 & 46.91 &	63.86	 \\ \hline
Baseline + ST-offsets &$\checkmark$  & &53.88 &	70.03	& 46.52&63.50 \\
Baseline + ST-offsets  & &$\checkmark$  &  59.15	 &74.33	 & 51.90	 &68.33	 \\ \hline
 Baseline + BT-flows  & $\checkmark$ & & 54.49	& 70.54  &47.37  &64.29	 \\
 Baseline  + BT-flows   & & $\checkmark$ & 58.90  &74.14 	& 52.14  &68.54 	 \\ 
\hline
\end{tabular}
\label{tab:abl_3}
\vspace{-3mm}
\end{table}

\pparagraph{Influence of the prediction method of the vector fields.} We compared the performance of the baseline model with the classification method and the regression method of the vector fields. The experimental results are illustrated in the first three rows of Tab.~\ref{tab:abl_3}. It shows that the classification method  brings noticeable gain in contrast to the regression method, which improves the change mask IoU on  the out-domain test set from 46.52$\%$ to 51.90$\%$. And the regression task of ST-offsets reduces the performance of baseline model by 0.39$\%$ in terms of IoU.  The same conclusion is also obtained for the BT-flows task. One possible reason for the above results is that the regression problems are more difficult to learn than classification problems. Our model implicitly uses offset features instead of the precise vector fields, So we choose the classification method of vector fields.We use the endpoint error (EPE) loss, which is commonly used in optical flow regression tasks, as the loss of regression single-temporal building offsets. It is the Euclidean distance between the predicted flow vector and the ground truth, averaged over all pixels.

\begin{table*}[!t]
\centering
\caption{Comparison of generalization cross dataset. $*$ means just use the part of  building change label in SECOND. }

\resizebox{1.0\linewidth}{!}{
\begin{tabular} {c|c|lllll|cc}
\hline
Metrics & \diagbox{Train on:}{Test on:} & BANDON & SECOND & S2Looking& LEVIR-CD & WHU-CD  &  \makecell[c]{Mean \\ drop }  & \makecell[c]{Percent \\ drop}  \\  \hline 
\multirow{4}*{IoU} & Off-Nadir-CD& \textbf{52.76 }& 56.69(-9.44) & 31.90(-15.12) & 40.69(-41.42) & 54.75(-32.06) & \textbf{24.51} & \textbf{33.45$\%$} \\
& SECOND$^{\bf*}$~\cite{yang2021asymmetric}& 39.25(-13.51) &\textbf{ 66.13} & 31.41(-15.61) & 47.62(-34.49) & 48.04(-38.77) & 25.60 & 36.37$\%$ \\
& S2Looking~\cite{shen2021s2looking}& 20.85(-31.91) & 41.11(-25.02) &\textbf{ 47.02} & 39.65(-42.46) & 44.04(-42.77) & 35.54 & 49.82$\%$ \\
& LEVIR-CD~\cite{chen2020spatial}& 9.01(-43.75) & 31.86(-34.27) & 3.51(-43.51) & \textbf{82.11} & 71.11(-15.70) & 34.31 & 61.34$\%$ \\
& WHU-CD~\cite{ji2018fully}& 8.74(-44.02) & 35.40(-30.73) & 7.06(-39.96) & 54.94(-27.17) & \textbf{86.81} & 35.47 & 61.99$\%$ \\
\hline
\multirow{4}*{F1} & Off-Nadir-CD&\textbf{ 69.07} & 72.36(-7.25) & 48.37(-15.60) & 57.84(-32.33) & 70.76(-22.18) & 19.34 & 23.30$\%$ \\
&  SECOND$^{\bf*}$~\cite{yang2021asymmetric}& 56.37(-12.70) &\textbf{ 79.61} & 47.80(-16.17) & 64.52(-25.65) & 64.90(-28.04) & 20.64 & 25.57$\%$ \\
& S2Looking~\cite{shen2021s2looking}& 34.51(-34.56) & 58.27(-21.34) & \textbf{63.97} & 56.79(-33.38) & 61.15(-31.79) & 30.27 & 37.02$\%$ \\
& LEVIR-CD~\cite{chen2020spatial}& 16.54(-52.53) & 48.32(-31.29) & 6.79(-57.18) & \textbf{90.17} & 83.12(-9.82) & 37.70 & 53.83$\%$ \\
& WHU-CD~\cite{ji2018fully}& 16.07(-53.00) & 52.29(-27.32) & 13.19(-50.78) & 70.92(-19.25) & \textbf{92.94} & 37.59 & 52.95$\%$ \\

\hline
\end{tabular}}
\label{tab:cross}
\vspace{-2mm}
\end{table*}

\subsection{Quality Analysis of the BANDON Dataset}
\label{sec:BANDON_ANY}
 BANDON contains a large number of off-nadir aerial images, and has many characteristics which are more similar to those of real-world images than that in the existing BCD datasets, \ie, large differences in image radiations, diverse types of changes, and concurrence of urban and rural scenes.  \\
In order to verify the above advantages of Off-Naidr-CD, we used the ChangeFormer~\cite{daudt2018fully} with SegFormer-b2 backbone to perform generalization experiments between the four datasets, including BANDON, WHU-CD~\cite{ji2018fully}, LEVIR-CD~\cite{chen2020spatial}, S2Looking~\cite{shen2021s2looking} and SECOND~\cite{yang2021asymmetric}.  In order to maintain a consistent experimental setting, we cut the WHU-CD dataset into 1950 pairs of samples, and divide the training, validation, and test sets according to 8:1:1. The image size of the other three datasets remains the same as shown in Tab.\ref{tab:datasets}, and the division adopts their respective official schemes. The training, validation, and test sets of LEVIR-CD are 445/64/128,  the training and test sets of S2looking are 3500/1000, and the training and test sets of SECOND are 2968/1694.
SECOND is a semantic change detection data set, so we only use its building annotations for experiments. The cross-dataset experimental results are shown in Tab.~\ref{tab:cross}. Each row corresponds to training on one dataset and testing on all the others. It proves that our dataset has the best generalization on both metrics, which is mainly due to the large-scale and diverse changing scenarios. So the BANDON is a suitable benchmark dataset for building change detection,  and the creation of the BANDON dataset is very meaningful for research in this field.

\subsection{Results and Analysis on Other Existing Datasets}
In order to further explore and analyze the effect of each component in our method, We again conducted further validation experiments using an existing public dataset. It can be seen from Section \ref{sec:BANDON_ANY} and Tab.~\ref{tab:datasets} that the SECOND dataset has sufficient data and covers a wide variety of scenarios, which makes the experimental conclusions on it more convincing. So we conduct additional experiments on the SECOND dataset and show the results with other representative methods in Tab.~\ref{tab:second_sota}. The training schedule is 25000 iterations, and other 
implementation details are consistent with Section~\ref{sec:exp_details}. The SECOND dataset has only additional segmentation labels, so our method is implemented with only the segmentation auxiliary task.  Nevertheless, our method still performs better than other competitors without auxiliary or partial auxiliary tasks, both in terms of IoU and $F_{1}$-score. Moreover, this result also proves that our multi-task learning method, including MTFGM, is also effective on near-nadir images.

\begin{table}[!t]
\vspace{-0mm}
\footnotesize

\begin{center}
\caption{Quantitative comparison with state-of-the-art  methods on the SECOND test dataset. }
\vspace{-0mm}
\setlength\tabcolsep{13pt}
\begin{tabular} {c|c|cccc}
\hline
 Method & Backbone & IoU & $F_1$ & Recall & Precision \\ \hline
 FC-siam-Diff\cite{daudt2018fully} & ResNet-50 & 62.91	& 77.23 &	76.18 &	78.32\\
 FC-Siam-Conc\cite{daudt2018fully} & ResNet-50 &  63.14 &	77.41 &	76.06 &	78.81\\
 DTCDSCN\cite{liu2020building} & SE-ResNet-34  & 62.52 &	76.94 &	76.47&	77.40	 \\
 Change-Mix\cite{zheng2021change} & ResNet-50 &  63.18	& 77.44 &	72.57 &	\textbf{83.00} \\ 
 ChangeFormer~\cite{bandara2022transformer} & SegFormer-b2 &  66.13 & 79.61 & \textbf{ 78.79} & 80.45\\
\hline
 Ours & ResNet-50 & \textbf{66.81} & \textbf{80.11}	& 78.01 & 82.32 \\ \hline
\end{tabular}
\label{tab:second_sota}
\end{center}
\vspace{-4mm}
\end{table}

\subsection{Limitations}
Although the MTGCD-Net model can meet the challenges of building change detection in off-nadir aerial images, the annotations required for training are non easy to obtain. Semi-supervised or unsupervised methods need to be introduced to migrate MTGCD-Net to data without auxiliary labels for training, such as MuST~\cite{ghiasi2021multi}, \etc. We leave it as future work.

\section{Conclusion}
In this paper, we have addressed the problem of building change detection with off-nadir aerial image pairs. A new MTGCD-Net model is proposed to tackle the challenging off-nadir building change detection task. The carefully designed auxiliary tasks of MTGCD-Net provide indispensable building parsing and matching information, which greatly helps to improve the performance of the main building change detection task.  A new dataset, named BANDON, is also created to train and evaluate building change detection models for Off-nadir aerial image pairs. Extensive experiments on the newly built dataset BANDON demonstrate the effectiveness and superiority of our method, comparing with previous methods.

\Acknowledgements{This work was supported in part by National Nature Science Foundation of China (Grant No. 41820104006, U22B2011, 61922065), and National Key R$ \& $D Program of China (Grant No. 2021YFB3900503).}





\end{document}